\documentclass[10pt]{cai}
\graphicspath{{figs/}}

\usepackage{adjustbox}
\usepackage{makecell}
\usepackage{amsmath,amssymb,amsfonts}
\usepackage{graphicx}
\usepackage{textcomp}
\usepackage{xcolor}
\usepackage{multirow}
\usepackage{caption}
\usepackage{subcaption}
\begin{document}

\newtheorem{definition}{Definition}
\def\conferenceyear{2024}
\volumeheader{37}{0}
\begin{center}

\title{Edge-DIRECT: A Deep Reinforcement Learning-based Method
for Solving Heterogeneous Electric Vehicle Routing Problem with
Time Window Constraints}
\maketitle

\thispagestyle{empty}

\begin{tabular}{cc}
      Arash Mozhdehi\upstairs{\affilone}, Mahdi Mohammadizadeh\upstairs{\affilone} , Xin Wang\upstairs{\affilone*}
   \\[0.25ex]
   {\small \upstairs{\affilone} University of Calgary, Calgary, AB, Canada} \\
\emails{
  \upstairs{*}xcwang@ucalgary.ca
}
\end{tabular}
  
\vspace*{0.2in}
\end{center}

\begin{abstract}
In response to carbon-neutral policies in developed countries, electric vehicles route optimization has gained importance for logistics companies. With the increasing focus on customer expectations and the shift towards more customer-oriented business models, the integration of delivery time-windows has become essential in logistics operations. Recognizing the critical nature of these developments, this article studies the heterogeneous electric vehicle routing problem with time-window constraints (HEVRPTW). 
To solve this variant of vehicle routing problem (VRP), we propose a  DRL-based approach, named Edge-enhanced Dual attentIon encoderR and feature-EnhanCed dual aTtention decoder (Edge-DIRECT). Edge-DIRECT features an extra graph representation, the node connectivity of which is based on the overlap of customer time-windows. Edge-DIRECT's self-attention encoding mechanism is enhanced by exploiting the energy consumption and travel time between the locations. To effectively account for the heterogeneity of the EVs' fleet, a dual attention decoder has been introduced. Experimental results based on two real-world datasets reveal that Edge-DIRECT outperforms a state-of-the-art DRL-based method and a well-established heuristic approach in solution quality and execution time. Furthermore, it exhibits competitive performance when compared to another leading heuristic method.
\end{abstract}

\copyrightnotice

\section{Introduction}
\label{introduction}
Recently, electric vehicles (EVs) have surged in popularity, driven by global commitments to reduce carbon emissions and increased environmental awareness among corporations \cite{lin2016electric, waraich2013plug, lu2020time}. This trend is particularly evident in the logistics sector, where companies are actively integrating EVs into their transportation fleets. At the heart of this transition is the electric vehicle routing problem (EVRP), an optimization problem central to the operations of these logistics companies, focusing on dealing with the complexities of deploying EVs instead of internal combustion engine vehicles. This article addresses a practical routing problem for EVs, named heterogeneous electric vehicle routing problem with time-window constraints (HEVRPTW). It considers both vehicle attributes, such as varying cargo and battery capacities \cite{ghannadpour2019multi} and customer preferences regarding delivery times \cite{tacs2014vehicle}. These factors create a more realistic and applicable model for contemporary logistics challenges.
HEVRPTW, recognized as an NP-hard optimization problem, seeks to determine a set of routes with minimal cost, total traveling time, or total traveling distance, for a fleet of Heterogeneous EVs to serve each geographically dispersed customer's demands within a specified time-window.

Traditional methods, including exact and heuristics solvers, are conventionally employed to solve various vehicle routing problem (VRP) variants. Due to the NP-Hard nature of HEVRPTW, and VRPs in general, exact methods, such as branch-and-price \cite{wu2023branch} and branch-and-price-and-cut \cite{duman2022branch}, consume prohibitively long time for solving practical-size problems \cite{duan2020efficiently}. Heuristics solvers, on the other hand, are significantly faster than exact methods, trading optimal solutions for speed. However, these approximate methods
rely on human expertise and domain knowledge, leaving room for improvement \cite{jingwen2022deep}. With their success in effectively solving various combinatorial optimization problems, deep reinforcement learning (DRL)-based models with Transformer-style policy networks have been applied to different VRP variants, yielding similarly promising results. 
Despite the wide spectrum of VRPs being tackled with these DRL-based methods, the HEVRPTW variant has been left unstudied, and they fail to effectively solve this routing problem due to the three major limitations.
Firstly, these approaches fail to consider the reachability of nodes based on time-windows during the encoding phase, neglecting to account for how these temporal constraints influence the graph's structure. This negligence undermines their policy model's capability to distinguish between nodes that are feasibly reachable and those that are not, resulting in poor routing performance when there are time-windows associated with the deliveries. 
However, by harnessing the inherent structure of the problem, which factors in the reachability of nodes, we can enhance the search mechanism for generating feasible routes. This approach leads to improved routing performance when taking the time-window constraints into account. Secondly, they fail to capture the heterogeneous features of the vehicles concerning their battery and cargo capacity and to utilize this information in decision-making. The guidance of this state information can play an important role in a better policy search due to the disparity of the contribution of vehicles in serving customers. Finally, another of their shortfalls is focusing only on travel time or distance while failing to consider the energy consumption associated with travel. However, given the limited battery capacity of EVs and their need for multiple recharges in a planning horizon, including this aspect is essential. In fact, it plays a crucial role in the overall traveling cost, i.e. travel time and distance.

Aiming to solve the HEVRPTW we propose a novel DRL-based method with a Transformer-style policy network, named Edge-enhanced Dual attentIon encoderR and feature-EnhanCed dual aTtention decoder (Edge-DIRECT). 
The main contributions of this article are summarized as follows:
\begin{enumerate}
    \item We introduce a graph-based representation where the customers are denoted as nodes with edges indicating overlapping time-windows, determined by the travel time between them. A graph attention layer has been utilized to capture the relationship between the neighboring nodes. By exploiting the underlying structure of the graph, the model is able to improve the routing performance for feasible route conduction based on time-window constituents associated with deliveries.   
    \item We propose a decoder that is comprised of two attention-based decoding modules to handle the heterogeneity of EVs concerning their battery and cargo capacity by directly indexing a vehicle in a fleet and capturing the feature heterogeneity of the vehicles. 
    \item We enhance both the encoder and the decoder to effectively capture the energy consumption between locations and utilize it for improving routing performance. An edge-enhanced encoder has been introduced to embed the energy consumption associated with traveling between the nodes into the problem representation. The decoder utilizes the edge-enhanced problem representation along with the energy state of the heterogeneous vehicles for effective route construction for the EV fleet.

    \item We evaluate Edge-DIRECT's performance through experiments on instances derived from real-world traffic data in two major Canadian cities, Edmonton and Calgary. The findings reveal that Edge-DIRECT outperforms baselines—a conventional heuristic and state-of-the-art DRL-based method—in solution quality. Additionally, it competes favorably with another advanced heuristic method and achieves more rapid routing times than all compared baselines.

\end{enumerate}
The remainder of this article is structured as follows. Section \ref{section:relatedworks} gives a brief overview of the related works. In Section \ref{section:definition}, we formally define the HEVRPTW. Section \ref{section:methodology} details the Edge-DIRECT's framework. In Section \ref{section:experiments}, the computational experiments and analysis are presented. Finally, this article is concluded and future research directions are presented in Section \ref{section:conclusion}.

\section{Related Works}
\label{section:relatedworks}
In the operations research literature, only a few studies have focused on HEVRPTW. We briefly review these and then explore DRL-based approaches proposed for solving different VRPs.

The study by Hiermann et. al. \cite{hiermann2013electric} introduced an adaptive large neighborhood search-based heuristic method to tackle the HEVRPTW. In another study, Sassi et. al. \cite{sassi2014vehicle} presented a mixed integer programming model formulation for HEVRPTW and proposed a constructive-based heuristic method combined with a local search heuristic using an inject-eject technique for solving this NP-hard optimization problem. While numerous studies have employed heuristic solvers for VRPs, a significant drawback of these methods is their reliance on complex, manually-engineered search rules. Secondly, these methods suffer from a lack of generalizability and adaptability. Even a minor modification to the problem may necessitate rerunning the algorithm to find a solution \cite{li2020deep}.

Motivated by promising results utilizing DRL-based methods for solving combinatorial optimizations, Nazari et al. \cite{nazari2018reinforcement} proposed a DRL algorithm for solving VRP with an RNN-based decoder for its policy model and utilized the aggregated Euclidean travel distance as feedback for training the model. 
In subsequent research, Kool et al. \cite{kool2018attention} enhanced the Deep Reinforcement Learning (DRL)-based method by utilizing a Transformer-based policy network. They also employed a self-critical REINFORCE algorithm for training the policy model, which resulted in performance surpassing competitive baselines, including Google OR-Tools. Duan et al. \cite{duan2020efficiently} argued that previously proposed methods fall short in effectively solving the VRP when actual travel distances replace aggregated Euclidean distances. In such scenarios, these methods fail to outperform OR-Tools. To address this, they proposed using a graph convolutional network that leverages the actual travel distances between locations, successfully surpassing the performance of OR-Tools in terms of total travel distance. However, none of these studies investigated the EVRP were vehicles due to their limited battery capacity need to be recharged multiple times within a planning horizon. The study by Lin et. al. \cite{lin2021deep} tried to improve the previous DRL-based methods for solving the EVERP with time-window constraints. To this end, Structure2Vec is used to capture the EVs features as well as the location and demands of the customers. A decoder comprised of an attention-based model and an LSTM-based was used for route construction. However, this method utilizes a complete graph to represent and embed the problem using its encoder. Hence, it fails to consider the reachability of the node based on their time-window constants and captures the underlying structure for effective routing when time-windows are considered. Besides, the RNN-based model used in the decoder leads to insufficient parallel processing and reduced computational efficiency compared to decoder models used by previous studies such as  Kool et al. \cite{kool2018attention}. Thirdly, this proposed method does not consider the heterogeneity of the fleets in real-world logistics operations. Finally, this approach does not consider utilizing the energy consumption between the locations and only relies on the travel distance for decision-making despite its importance for effective policy search considering the limited capacity of EV batteries.

\section{Problem Definitions and Formulation}
\label{section:definition}
In this section, we define the HEVRPTW and express this routing problem through Markov decision process (MDP) formulation. 

\begin{definition} \label{def:twgraph}
\textbf{(Time-Window Graph).} The time-window graph is a directed graph defined as \( G = (N, E) \), where \( N = \{n_0, \ldots, n_{\mathcal{C} + \mathcal{E}}\} \) represents a set of nodes, including \( \mathcal{C} \) customer nodes, $\mathcal{E}$ charging stations, a depot $n_0$. For each node \( n_i \), a tuple \( (x_{i}, y_{i}, tw^1_{i}, tw^2_{i}) \) is assigned, where \( x_{i} \) and \( y_{i} \) are the 2-dimensional coordinates of the node \( n_i \), and \( tw^1_{i} \) and \( tw^2_{i} \) indicate the earliest and latest times, respectively, that a customer \( n_i \) (with \( 1 \leq i \leq \mathcal{C} \)) accepts delivery. Notably, for the depot and charging station, the values of \( tw^1_{i} \) and \( tw^2_{i} \), \( i \in {n_0} \cup N_e \) are set to 0 and \( T_{max} \), respectively. A customer set \( N_c \) identified as \( N_c = \{n_1, \dots, n_{C}\} \) is defined to stand for the set of nodes representing the customers. 
\( N_e \) denotes the set of nodes standing for the charging stations.
The time window graph's edge set \( E \) is denoted by \( E = \{e_{ij} \mid 0 \leq i, j \leq \mathcal{C} + \mathcal{E}, i \neq j\} \).  $\forall e_{ij} \in E$, we assign a tuple $(ec_{ij}, tt_{ij})$, where $ec_{ij}$ represents the energy required and $tt_{ij}$ denotes the travel time from nodes $n_i$ to $n_j$. 
The adjacency matrix for this time-window graph is represented as \( A \in \mathbb{R}^{(\mathcal{C} + \mathcal{E})\times (\mathcal{C} + \mathcal{E})} \), where each element \( a_{ij} \) is set to 1 if and only if $\exists x \in [tw^1_{i}, tw^2_{i}]$ where $x + tt_{ij} \in [tw^1_{j}, tw^2_{j}]$ and 0 otherwise.

\begin{definition} \label{def:dgraph}
\textbf{(Demand Vector).} Given the time-window graph's node set $N$, a demand vector $\mathcal{D} = \{d_{i} | i \in N_c\}$ is defined, where  \( d_{i} \) indicates the demand of the node, which is 0 when $i=0$ or $i > \mathcal{C}$.
\end{definition}

\end{definition}

\begin{definition} \label{def:fleet}
\textbf{(Vehicle Fleet Feature Vector).} A vehicle fleet feature vector is expressed as $\mathcal{F} = \{f_1, \dots, f_\nu\}$, where $\nu$ denotes the number of EVs in the fleet. Each element $f_j$ is a tuple $(\mathcal{E}_j, Q_j)$, where $\mathcal{E}_j$, $Q_j$ are respectively denoting the battery and cargo capacity of the vehicle $j$. 
\end{definition}

Tackling the HEVRPTW in a constructive manner is akin to a sequential decision-making problem.  This approach is formulated as an MDP, characterized by a quartet $\{\mathcal{S}, \mathcal{A}, \mathcal{P}, r\}$ which symbolizes the state space, action space, transition probabilities, and reward function, respectively. The detailed definitions of each component are outlined below:

\begin{itemize}
  \item \textbf{State:} 
  $s_T=(s^{F}_T,s^{R}_T) \in \mathcal{S}$ represents the system state at the step $T$, 
where $s^{\mathcal{F}}_T = \left[s^{{f}_1}_T, \dots, s^{{f}_\nu}_T\right]$, denotes the current state of the vehicles' fleet. The $i$-th element of vehicles' state is defined as $s^{{f}_i}_T=\left[rc^i_T, \mathcal{N}^i_T, re^i_T\right]$. Here, $rc^i_T$ and $\mathcal{N}^i_T$ respectively denote the vehicle $i$'s remaining capacity and the currently visited node. $re_T$ represents the remaining energy of the vehicle $i$. The second state element, $s^R_{T}$, refers to the routing state. For step $T$, this is conveyed through $s^{R}_T=\left\{N_T^{visited}, N_T^{to - visit}\right\}$. $N_T^{visited}$ includes all customer nodes visited so far, while $N_T^{to - visit}$ lists the customer nodes yet to be served.
  \item \textbf{Action:} $a_T \in \mathcal{A}$ represented as $(n_i, f_j)$ the is action at step $T$, where $a_T$ signifies that $n_i$ is visited by vehicle $f_j$ at step $T$. 
  \item \textbf{Transition:} 
  The state transition function is denoted as $s_{T+1}=\mathcal{P}(s_T, a_T)$, transitioning the system from its current state $s_T$ to the subsequent state $s_{T+1}$ in response to the most recent action $a_T$.

  \item \textbf{Reward:}
  Considering that HEVRPTW seeks to minimize the cumulative travel time of the vehicles, at any given decision step $T$, the reward for that step is established by $r(s_{T},a_{T}) = -tt_{ij}$, where $tt_{ij}$ denotes the travel time from node $n_i$ to node $n_j$.
\end{itemize}
\section{Methodology}
\label{section:methodology}
In this section, we introduce our deep reinforcement learning (DRL)-based method, termed Edge-enhanced Dual attentIon encoderR and feature-EnhanCed dual aTtention decoder (Edge-DIRECT), designed for tackling the HEVRPTW. 
We first delve into the Transformer-style policy network of this method. Then, we describe the DRL algorithm employed to train this policy network.

\begin{figure}[h!]
    \includegraphics[width=5.5in]{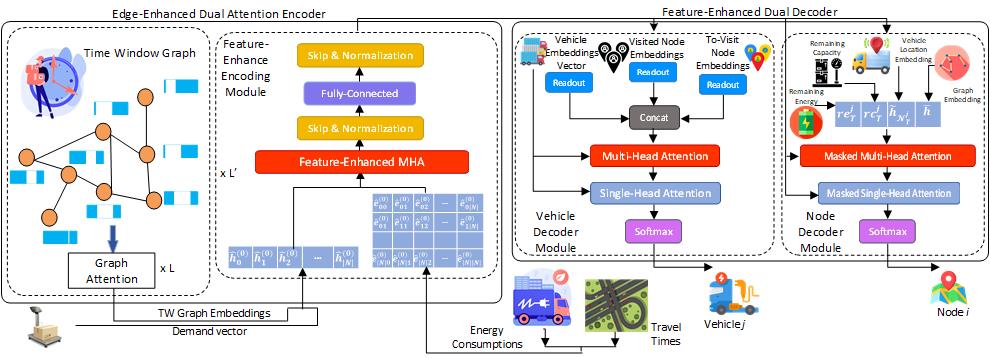}%
  
  \caption{Illustration of Edge-DIRECT's policy network architecture.}
  \label{diagram:framwork}
\end{figure}

\subsection{Policy Network Architecture of Edge-DIRECT}

Edge-DIRECT, as illustrated in Figure \ref{diagram:framwork}, employs a Transformer-style policy network to solve the HEVRPTW instance constructively, which involves sequentially adding a node to the solution at each decoding step. The policy network of Edge-DIRECT comprises two primary elements: an edge-enhanced dual attention encoder and a feature-enhanced dual decoding module. The functionalities and characteristics of each of these components are detailed in the subsequent sections.

\subsubsection{Edge-Enhanced Dual Attention Encoder}

The edge-enhanced dual attention Encoder projects the raw features of an instance into a higher-dimensional space, facilitating the extraction of features pertinent to the problem instance.
This encoder consists of two key encoding modules: a time-window graph attention module and a feature-enhance self-attention encoding module.

In the time-window graph attention module, given the time-window graph \(G = (N, E)\), a \(L\)-layer graph attention (GAT) operation, as detailed in \cite{velivckovic2017graph}, is applied. This method is specifically tailored to capture the correlations between neighbor nodes that are able to be reached from each other based on their time-window constraints. The introduction of a time-window graph, coupled with this technique for node correlation extraction, is to enhance the policy network's decision-making process, in the presence of time-window constraints, for better routing performance.

In each layer \( l \), for every element \( a_{ij} \) in the time-window graph's adjacency matrix, an attention coefficient \( \alpha_{ij}^{(l)} \) is calculated at the \( l \)-th GAT layer. This coefficient, determined by a learnable weight vector \( \mathbf{a}^{(l)} \) and a learnable weight matrix \( \mathbf{W}^{(l)} \), quantifies the influence of neighboring nodes, as follows:
\begin{eqnarray} \label{eq:gatcoeff}
\alpha_{ij}^{(l)} = \frac{\exp(\text{ReLU}(\mathbf{a}^{(l)T}[\mathbf{W}^{(l)}h_i^{(l-1)} \| \mathbf{W}^{(l)}h_j^{(l-1)}]))}{\sum_{k \in \mathcal{N}(i)} \exp(\text{ReLU}(\mathbf{a}^{(l)T}[\mathbf{W}^{(l)}h_i^{(l-1)} \| \mathbf{W}^{(l)}h_k^{(l-1)}]))} \cdot a_{ij}
\end{eqnarray}
where \( \| \) denotes the concatenation operation, ReLU represents the ReLU non-linear activation function, and \( h_i^{(l-1)} \) is the feature vector of the node \( n_i \) at layer \( (l-1) \). For layer 0, this feature vector is equivalent to a linear projection of the attributes associated with node \( n_i \), as described in Definition \ref{def:twgraph}. 

Then, the feature update rule for each node in the \( l \)-th GAT layer is calculated ${h^\prime}_i^{(l)} = \text{ReLU}\left(\sum_{j \in \mathcal{N}(i)} \alpha_{ij}^{(l)}\mathbf{W}^{(l)}h_j^{(l-1)} \right)$,

where \( {h^\prime}_i^{(l)} \) denotes the updated features of node \( n_i \) at the \( l \)-th layer. 

After completing the \( L \) layers of the Graph Attention Network (GAT), the output from the final encoding stage for each node \( n^{TW}_i \) is represented by \( {h^\prime}_i^{(L)} \). For the sake of brevity and clarity in the remainder of this article, we will refer to this final layer output for each node simply as \( h^\prime_i \).

In the feature-enhance self-attention encoding module, the process of embedding the problem instance into a high-dimensional representation is performed by enhancing the Transformer-based encoder presented in the works of Kool et al. \cite{kool2018attention} and Li et al. \cite{li2021heterogeneous}. For efficient decision-making based on energy consumption patterns and actual travel time between nodes, it is crucial that the policy model incorporates this information in its routing process. This can be achieved by capturing this information when the encoder embeds the problem into a high-dimensional representation. Given that these two factors depend on both the source and destination nodes, we propose an edge-enhanced encoder design. This design effectively captures the energy consumed and travel time as edge features in the time window graph, in addition to the node features.

Given the demand vector $\mathcal{D}$ and the updated node features \( h^\prime_i \), $i \in N$, the feature \( \hat{h}_i \) attributed to each node \( n_i \) is defined as the tuple \( (h^\prime_i, d_{i}) \). An edge feature \( \hat{e}_{ij} \) is defined, representing the feature vector allocated to the edge \( e_{ij} \) within the time window graph, as described in Definition \ref{def:twgraph}. The embeddings are subjected to a linear projection and a batch normalization (BN), resulting in the initial embeddings for nodes and edges, denoted as \( \hat{h}^{(0)}_i \) and \( \hat{e}^{(0)}_{ij} \), respectively. 

These embeddings are processed through \( L^\prime \) layers of edge-enhanced attention, each comprising an edge-enhanced multi-head attention sub-layer followed by a fully connected sub-layer. 

At each layer $l'$, the operation performed in each multi-head attention sub-layer involves first computing an attention coefficient \( \alpha_{ij}^{(l, k)} \) for each head \( k \) as follows:   
\begin{eqnarray} \label{eq:gatsum}
 \alpha_{ij}^{(l', k)} = \frac{\exp\left(\text{ReLU}\left(\mathbf{W}^{(l', k)^T} \left[\hat{h}^{(l')}_i || \hat{h}^{(l')}_j || \hat{e}^{(l')}_{ij}\right] \right)\right)}{\sum_{m \in \mathcal{N}(i)} \exp\left(\text{ReLU}\left(\mathbf{W}^{(l', k)^T} \left[\hat{h}^{(l')}_i || \hat{h}^{(l')}_m || \hat{e}^{(l')}_{im}\right] \right)\right)} 
\end{eqnarray}
Here, \( \mathbf{W}^{(l', k)} \) denotes a learnable weight vector for each attention head $k$ in the layer $l'$. 

The next layer node layer representation  \( \hat{h}^{(l'+1)}_i \) is then calculated as follows:
\begin{eqnarray} \label{eq:gatsum}
\hat{h}^{(l'+1)}_i = BN\left(\hat{h}^{(l')}_j + \left(\text{ReLU}\left(\bigoplus_{k=1}^{K} \sum_{j \in \mathcal{N}(i)} \alpha_{ij}^{(l', k)} \mathbf{W}^{(l', k)} \hat{h}^{(l')}_j \right)\right)\right)
\end{eqnarray}
where $\bigoplus$ denotes the concatenation operation, and \( K \) represents the number of attention heads, and \( \mathbf{W}^{(l', k)} \) is a weight matrix for the \( k \)-th attention head.

Subsequently, the outputs of the \( l' \)-th layer are produced by processing the intermediate representations through a fully connected (FF) layer with a ReLU activation function. This sub-layer includes a skip connection \cite{he2016deep} and a batch normalization layers \cite{ioffe2015batch}. 

Finally, the encoder's output, denoted as \( \Tilde{h}^{(L')}_i \), is obtained after processing through the final sub-layer at the \( L' \)-th layer. For brevity, we will use \( \Tilde{h}_i \) in the rest of this article, omitting the superscript \( (L') \). A graph embedding \( \Bar{h} \) is then computed for each problem instance by averaging these node representations: \( \Bar{h} = \frac{1}{|N|}\sum_{i=0}^{|N|} \Tilde{h}^{(L')}_i \).
\subsubsection{Feature-Enhanced Dual Decoder}

Given node embeddings \( \tilde{h}_i \), where \( i \) ranges from 0 to \( |N| \), derived from the encoder, and the aggregate graph embedding \( \bar{h} \), the decoder, at each step \( T \), produces two distinct probability distributions: \( p^f_T \) for selecting a vehicle from a heterogeneous fleet, and \( p^n_T \) for determining the node to be visited by the chosen vehicle. This setup includes two dedicated modules: the vehicle decoder module and the node decoder module.

Each decoding step starts by selecting a vehicle from a heterogeneous fleet using the vehicle decoder module. This module allows the policy model to make informed routing decisions for a fleet of heterogeneous EVs, by capturing the diverse characteristics of these vehicles. For this vehicle decoding module, we propose employing a deep learning model with cross-attention mechanism. 

Given the vehicle fleet features vector $\mathcal{F}$ and fleet's state $s^{F}_T$, for each vehicle $j$, a high-dimensional representation ${h}^{f_j}_T$ is computed by: ${h}^{f_j}_T=FF\left[FF\left(f_j\right) ||  FF\left(s^{{f}_j}_T\right) \right]$

, where \( FF \) represents a two-layer, fully connected network with ReLU activation functions. It is important to note that for efficiency, the operation of the fully connected network on \( f_j \) is performed only at the start of each episode.

Then, given the vehicle embeddings vector ${H}^F_T = \left\{{h}^{f_1}_T, \dots, {h}^{f_\nu}_T\right\}$  and node embeddings $\Tilde{h}_i, i \in N$, a context embedding $h^{(c)^F}_T$ is computed through the following equation:
\begin{equation} \label{eq:gatsum}
h^{(c)^{F}}_T = \left[ \text{Readout}\left({H}^F_T\right) || \text{ Readout}\left(\bigcup_i \in N_T^{visited}\{\Tilde{h}_i\}\right)\} || \text{ Readout}\left(\bigcup_{i \in N_T^{to - visit}}\{\Tilde{h}_i\}\right) \right]
\end{equation}
where Readout is denotes the standard Readout operation \cite{buterez2022graph}.

Afterward, given the vehicle embeddings $\Bar{H}^F$ and the context embedding $h^{(c)^{F}}_T$, using a Multi-Head Attention (MHA) operation, as described in Kool et al. \cite{kool2018attention}, an embedding ${h^{(g)^F}_T} = MHA(W^{(g)^F}_Q h^{(c)^{F}}_T, W^{(g)^F}_K \Bar{H}^F, W^{(g)^F}_V \Bar{H}^F)$ is calculated. $W^{(g)^F}_Q$, $W^{(g)^F}_K$ and $W^{(g)^F}_V$ are respectively the learnable weights for the $query$, $key$, $value$. Using a Single-Head Attention \cite{vaswani2017attention}, and given the embedding ${h^{(g)^F}_T}$ and the vector $\bar{H}^F_T$, with learnable parameters $W^{(g)^F}_Q$, $W^{(g)^F}_K$, and $W^{(g)^F}_V$ for $query$, $key$, and $value$ respectively, the $compatibility$ ${h^{(c)^F}_T}$ is calculated as ${h^{(c)^F}_T} = SHA(W^{(c)^F}_Q h^{(g)^F}_T, W^{(c)^F}_K \bar{H}^F_T, W^{(c)^F}_V \bar{H}^F_T)$. 
Finally, a probability distribution \( p^f_T \) over the heterogeneous fleet is computed through a Softmax operation on ${h^{(c)^F}_T}$.

At the beginning of node decoding process, given the selected vehicle $j$, problem instance's graph embedding $\Bar{h}$, node representations $\Tilde{H}=\left\{\Tilde{h}_{0}, \dots, \Tilde{h}_{|N-1|}\right\}$, and vehicle state $s^{V}_T$, a context embedding $h^{(c)}_T$ is computed through: $h^{(c)}_T=\left[re^j_T || rc^j_T ||  \Tilde{h}_{\mathcal{N}^j_{T}}  ||  \Bar{h}\right]$

Here, \( \mathcal{N}_{T} \), \( re^j_T \), and \( rc^j_T \) represent the current location of the vehicle $j$, its remaining energy, and its remaining capacity, respectively.

This novel context embedding equips the policy network with essential state information needed to handle the constraints of HEVRPTW, facilitating informed and effective decision-making in solving the routing problem. 

Next, given the context embedding $h^{(c)}_T$ and encoded node features vector $\Bar{H}$ at the decoding step $T$, using a  MHA layer, glimpse ${h^{(g)}_T} = MHA(W^{(g)}_Q h^t_{(c)}, W^{(g)}_K \Bar{H}, W^{(g)}_V \Bar{H})$ is computed.

 $W^{(g)}_Q \in \mathbb{R}^{d_h \times d_Q}$, $W^{(g)}_K \in \mathbb{R}^{d_h \times d_K}$, and $W^{(g)}_V \in \mathbb{R}^{d_h \times d_V}$ are the trainable parameters corresponding to the $query$, $kay$, and $value$, respectively. For every node \( n_i \), given the node embedding $\Tilde{h}_{0}$ and the $glimpse$ ${h^{(g)}_T}$, a compatibility score is determined to indicate its pertinence to a query, i.e. the $glimpse$, as part of the attention mechanism. This score measure is detailed, as follows:
\begin{eqnarray} \label{eq:kqv_comp_decoder2_tanh}
{u}^T_{(c),i} = 
\begin{cases}
    C.tanh\left(\dfrac{{\left(W^{(u)}_Qh^{(g)}_T\right)}^\intercal \left(W^{(u)}_K\Tilde{h}_{i}\right)}{\sqrt{d_k}}\right),& \text{if } M_{{i},{T}} = 0 \\
    -\infty,& \text{otherwise}\\
\end{cases}
\end{eqnarray} 
where $C$ is a constant for controlling the entropy. $M_{{i},{T}}$ denotes the asking matrix corresponding to the node $n_i$ and defined 
as follows:
\begin{eqnarray} \label{eq:mask}
    M_{{i},{T}} = 
\begin{cases}
    1 & \text{if } v_i \in N_{T+1}^{visited} \\
    1 & \text{if } rc^j_T < d_{i}\\
    1 & \text{if } \mathcal{N}_{T-1} = n_0 \text{ and }  \mathcal{N}_{T} \in \{n_0, n_{|N|-1}\}\\
    1 & \text{if } \forall j \in N_e \text{, } re^j_T < EC(\mathcal{N}_{T}, v_i) + EC(n_i, n_{j})\\
    0              & \text{otherwise}
\end{cases}
\end{eqnarray}
The first masking rule ensures that already-served customers are not visited again, while the second rule prohibits visits to customers whose demand exceeds the vehicle's remaining capacity. The third rule prevents visiting the depots if in the previous decoding step the depot was visited. The battery capacity constraint is enforced by the fourth masking rule.

Finally, given the $glimpse$ ${u}^T_{(c),i}$, the portability of vising node $n_i$ at the decoding step $T$ is computed using a Softmax layer, where the probability corresponding to the masked nodes is set 0. Every time a charging station is visited by the vehicle $j$, its remaining energy, i.e. $re^j_T$, resets to $\mathcal{E}_j$.

\subsection{DRL Algorithm}
For training the Transformer-style policy network of Edge-DIRECT, the REINFORCE method \cite{williams1992simple} has been employed. This reinforcement learning algorithm applies a policy-gradient approach with a greedy roll-out baseline \cite{rennie2017self}. It uses the total travel time of routes constructed at the conclusion of each episode as a feedback mechanism for refining the policy model's parameters during training.

\section{Experiments}
\label{section:experiments}

This section presents thorough experimental analyses using two real-world datasets, derived from traffic data of two major Canadian cities, Edmonton and Calgary, to address the subsequent research questions:

\begin{itemize}
    \item RQ1: How does Edge-DIRECT perform compared to the state-of-the-art DRL-based and heuristics methods in solving the HEVRPTW in terms of the aggregated travel time and computational time?
    \item RQ2: How do the time-windows encoding module, edge-enhanced encoding module, and dual attention decoder contribute to improving solution quality for the HEVRPTW?
    \item RQ3: How do varying ratios of the number of charging stations to the number of customers impact the results?
\end{itemize}

We conducted training and testing experiments for DRL-based models on servers equipped with A100 GPUs with 40 GB memory size. Servers with 32 cores Intel(R) Xeon(R) Gold 6240R CPUs. In the DRL models, we configure the hidden dimension size at 128, establish 3 encoding layers, and set the number of attention heads to 8. The model is trained using the Adam optimizer with the learning rate of $10^{-3}$ with a batch size of 256.

Experiments are conducted on problem sets of three distinct sizes: 20, 50, and 100. For these sets, customer demands are uniformly sampled from integers within the range of 1 to 9.
Locations are uniformly sampled from Edmonton and Calgary in Canada, with travel speeds for each road segment determined using historical traffic data. For the time-windows allocated to serving customers, the start times are uniformly sampled from the interval [0, 720], and the duration of each time-window is sampled from the range [60, 180]. For problem sizes of 20, 50, and 100, the vehicle and charging station configurations are as follows: Size 20 includes 3 vehicles and 3 charging stations, each with cargo capacities of 20, 30, and 40, and battery capacities of 450, 500, and 550 kWh, respectively. Size 50 comprises 5 vehicles and 5 charging stations with cargo capacities of 20, 30, 40, 50, and 60, and battery capacities of 400, 450, 500, 550, and 600 kWh, respectively. For size 100, there are 11 vehicles and 11 charging stations, with cargo capacities increasing from 25 to 75 in 5-unit increments, and corresponding battery capacities ranging from 375 to 625 kWh in 25 kWh steps. The recharging and discharging rate formula is adopted from Shen et al. \cite{shen2022robust}. The models were trained on 512,000 problem instances and tested on 2,000 instances.

\subsection{Performance Analysis (RQ1)}

In evaluating the performance of Edge-DIRECT for solving the HEVRPTW, we benchmarked its performance against heuristic methods and a state-of-the-art Deep Reinforcement Learning (DRL)-based approach. Specifically, we adapted and enhanced the Ant Colony Optimization (ACO) method originally proposed by Mavrovouniotis et al. \cite{mavrovouniotis2018ant} for the EVRP by integrating techniques from Han et al. \cite{han2022solving} to tackle time-window constraints and fleet heterogeneity. Furthermore, we utilized and enhanced the Variable Neighborhood Search / Tabu Search (VNS/TS) heuristic, based on the work by Schneider et al. \cite{schneider2014electric}, for EVRP within the context of heterogeneous fleets and time-window considerations. Alongside these heuristics, Edge-DIRECT's performance was compared with a DRL-based method, named EVRPTW-RL, as described by Lin et al. \cite{lin2021deep}, which employs an encoder-decoder structure specifically designed for EVRP challenges. To ensure a fair comparison, as EVRPTW-RL cannot accommodate vehicle selection, vehicles are randomly selected at the end of each trip. HEVRPTW-X denotes problem instances with X customers.

For DRL-based methods, there are two decoding strategies: 1. The Greedy approach, where at each decoding phase, the decoder selects the vehicle and the node being visits by that vehicle with the highest probability as indicated by the Softmax layers. 2. The Sampling strategy, where 1280 routes are sampled from the distributions, outputted by vehicle decoder and node decoder modules, and the route with the highest reward is selected.

As perceivable through Table \ref{tbl:performance}, Edge-DIRECT with sampling-based decoding outperforms the state-of-the-art DRL-based method proposed for EVs' routing, i.e. EVRPTW-RL, with with decoding strategies in terms of cost. As the size of the problem increases the performance gap widens favoring Edge-DIRECT over EVRPTW-RL despite requiring much less computational times, due to EVRPTW-RL's step-wise encoding. 
Edge-DIRECT (Sampling) significantly surpasses the performance of ACO, a well-established heuristic methods on both datasets. This superiority is not only evident in terms of solution quality but also in computational efficiency and the gap increases as the problem size grows. Edge-DIRECT with greedy decoding strategy mandates least amount of running time compared to all baselines and is able to exceeds the performance of ACO for problem of size 100 on both datasets. While the sampling strategy of Edge-DIRECT demands more computational time compared to its greedy-based counterpart, it remains competitive in terms of total travel time when juxtaposed with heuristic methods such as VNS/TS. Moreover, it benefits from significantly reduced computational time compared to both heuristic approaches and EVRPTW-RL (Sampling).

\begin{table*}[t]
\begin{adjustbox}{width=1.\textwidth}
\centering
\begin{tabular}{@{}l|ccc|ccc|cccc@{}}
\hline
\multicolumn{1}{c|}{} & \multicolumn{3}{c}{} & \multicolumn{3}{c}{Calgary Dataset} & \multicolumn{3}{c}{} \\
\cline{2-10}
\multicolumn{1}{c|}{} & \multicolumn{3}{c|}{HFEVRPTW-20} & \multicolumn{3}{c|}{HFEVRPTW-50} & \multicolumn{3}{c}{HFEVRPTW-100} \\
Model & Cost & Gap & Time & Cost & Gap & Time & Cost & Gap & Time \\
\hline
ACO & 678.6 & 5.60\% & 7.4 min & 1405.9 & 15.08\% & 41.8 min & 3133.7 & 24.64\% & 2.8 h \\
VNS/TS & 642.6 & 0.00\% & 6.6 min & 1222.5 & 0.00\% & 54.4 min & 2514.2 & 0.00\% & 15.8 h \\
\hline
EVRPTW-RL (Greedy) & 755.1 & 17.50\% & 1.81 s & 1639.6 & 34.12\% & 9.68 s & 4000.1 & 59.1\% & 22.06 s \\
EVRPTW-RL (Sampling) & 689.2 & 7.25\% & 16.1 min & 1392.3  & 13.89\% & 1.6 h &  3230.7 & 28.49\% & 3.5 h \\
\hline
Edge-DIRECT (Greedy) (ours) & 704.2 & 9.58\% & 1.51 s & 1441.4 & 17.90\% & 4.34 s & 3113.6 & 23.84\% & 11.10 s \\
Edge-DIRECT (Sampling)(ours)  & 660.5 & 2.78\% & 5.6 min & 1273.4 & 4.16\% & 25.8 min & 2699.7 & 7.37\% & 1.8 h \\

\Xhline{1.5pt}
\end{tabular}
\end{adjustbox}
\begin{adjustbox}{width=1.\textwidth}
\begin{tabular}{@{}l|ccc|ccc|cccc@{}}
\multicolumn{1}{c|}{} & \multicolumn{3}{c}{} & \multicolumn{3}{c}{Edmonton Dataset} & \multicolumn{3}{c}{} \\
\cline{2-10}
\multicolumn{1}{c|}{} & \multicolumn{3}{c|}{HFEVRPTW-20} & \multicolumn{3}{c|}{HFEVRPTW-50} & \multicolumn{3}{c}{HFEVRPTW-100} \\
Model & Cost & Gap & Time & Cost & Gap & Time & Cost & Gap & Time \\
\hline
ACO & 486.4 & 5.34\% & 16.3 min & 1208.3 & 16.42\% & 48.1 min & 2478.5 & 23.96\% & 2.6 h \\
VNS/TS & 461.7 & 0.00\% & 14.1 min & 1037.8 & 0.00\% & 56.2 min & 1999.3 & 0.00\% & 16.1 h \\
\hline
EVRPTW-RL (Greedy) & 537.6 & 16.43\% &  1.76 s & 1330.6 & 33.21\% & 9.82 s & 3213.9 & 60.75\% & 23.7 s \\
EVRPTW-RL (Sampling) & 491.1 & 6.36\% & 17.1 min & 1177.8 & 14.49\% & 1.7 h & 2641.3 & 32.11\% & 3.5 h \\
\hline
Edge-DIRECT (Greedy) (ours) & 506.8 & 9.76\% & 1.48 s & 1212.5 & 16.83\% & 4.58 s & 2419.6 & 21.02\% & 11.47 s \\
Edge-DIRECT (Sampling) (ours) & 479.5 & 2.94\% & 5.3 min & 1081.3 & 4.19\% & 25.1 min & 2138.1 & 6.94\% & 1.7 h \\
\hline
\end{tabular}
\end{adjustbox}
\caption{Edge-DIRECT vs. Baselines for Solving HFEVRPTW}
\label{tbl:performance}
\end{table*}

\subsection{Ablation Study (RQ2)}
Edge-DIRECT is featured with 3 major enhancements to effectively solve the HEVRPTW. The contribution of each of these features has been investigated on the Calgary Dataset with greedy decoding strategy and results are detailed in Table \ref{tbl:ablation}. "Edge-DIRECT w/o EE" represents Edge-DIRECT without the edge-enhanced encoder, wherein the encoder only exploits the node features. "Edge-DIRECT w/o TWE" indicates the variant lacking the time-window graph encoding module, where time-windows are instead assigned as raw features to each node. Finally, "Edge-DIRECT w/o HD" denotes the model variant excluding the specialized decoder designed to manage fleet heterogeneity.

Results demonstrate that the original Edge-DIRECT has better performance compared to the other variants. While removing the time-window encoding module notably reduces running time more than the other two variants, this module also yields a more substantial improvement in cost compared to the 2 enhancements to Edge-DIRECT. After the time window encoder, the addition of the heterogeneous vehicle decoder notably boosts cost efficiency. In terms of computational time, its impact is comparable to that of the time window encoder, yet marginally less. The edge-enhanced encoder contributes the least to cost reduction compared to the other modules and incurs minimal computational time among them.

\begin{table}[ht]
\centering
\begin{tabular}{@{}l|cc|cc|cc@{}}
\hline
\multicolumn{1}{c|}{} & \multicolumn{2}{c|}{HEVRPTW-20} & \multicolumn{2}{c|}{HEVRPTW-50} & \multicolumn{2}{c}{HEVRPTW-100} \\
Model & Cost & Time  & Cost & Time  & Cost & Time  \\
\hline
Edge-DIRECT w/o EE & 718.4 & 1.29 s & 1385.5 & 3.64 s & 3485.3 & 9.31 s  \\
Edge-DIRECT w/o TWE &  742.1 & 1.03 s  &  1449.3 & 3.08 s  & 3278.7 & 7.73 s \\
Edge-DIRECT w/o HD &  738.5 & 1.13 s  &  1445.5 & 3.22 s  & 3296.7 & 8.02 s  \\
Edge-DIRECT & 704.2 & 1.51 s & 1441.4 & 4.34 s & 3113.6 & 11.10 s \\ 
\hline
\end{tabular}
\caption{Edge-DIRECT's Ablation Study Results}
\label{tbl:ablation}
\end{table}

\subsection{Impact of Charging Station-to-Customer Ratio (RQ3)}
Experiments were conducted to explore the effect of varying the ratio of charging stations to customers on the performance of DRL-based methods, on the Calgary dataset for a problem instance comprising 100 customers. The findings are detailed in Table \ref{tbl:ratio}. Notably, increase in charging stations leads to a reduction in overall cost but a rise in computational time, attributable to the enhanced complexity from incorporating more nodes into the encoding and decoding processes. Moreover, as the number of charging stations escalates, the advantage in total travel time progressively shifts in favor of Edge-DIRECT over EVRPTW-RL, underscoring Edge-DIRECT's superior capability in leveraging charging stations to optimize the total travel cost effectively.
\begin{table}[ht]
\centering
\begin{tabular}{@{}l|cc|cc|cc|cc@{}}
\hline
 & \multicolumn{2}{c|}{2\%} & \multicolumn{2}{c|}{5\%} & \multicolumn{2}{c|}{10\%} & \multicolumn{2}{c}{20\%} \\
Model & Cost & Time & Cost & Time & Cost & Time & Cost & Time \\
\hline
EVRPTW-RL & 4024.1 & 15.78 s & 4018.1 & 15.83 s & 4000.1 & 15.98 s & 3987.4 & 17.31 s \\
Edge-DIRECT (ours) & 3201.5 & 8.18 s & 3165.2 & 8.24 s & 3113.6 & 8.32 s & 3068.3 & 8.65 s \\
\hline
\end{tabular}
\caption{Comparison of The Impact of Different Station-to-Customer Ratios}
\label{tbl:ratio}
\end{table}

\section{Conclusion}
In this study, we address the HEVRPTW by introducing a DRL-based method, Edge-DIRECT. This approach features a Transformer-style policy network with an edge-enhanced encoder that leverages energy and travel costs between nodes for efficient EV routing. To accommodate time-window constraints, Edge-DIRECT employs a GAT layer to discern the graph's structure from the nodes' connectivity based on time-window constraints. Additionally, a dual-attention decoder is utilized to account for fleet heterogeneity, focusing on battery and cargo capacity. Through comprehensive testing on datasets from Edmonton and Calgary, Edge-DIRECT's efficacy is assessed against leading heuristics and state-of-the-art DRL methods. 
The results demonstrate that Edge-DIRECT outperforms EVRPTW-RL, a competitive deep reinforcement learning (DRL)-based approach designed for addressing Electric Vehicle Routing Problems with Time-Windows (EVRPTW), in terms of solution quality with significantly reduced computational time.
Notably, it competes well with the VNS/TS and surpasses ACO in route efficiency, with substantially less running time.

There are several avenues for future research endeavors. First, in this article we assumed that the energy recharging and consumption is linear.  However, we aim to delve into more complex, non-linear models of consumption and recharging in subsequent research, aligning our approach more closely with real-world dynamics. 
Second, we plan to explore routing variations that accommodate the potential for order cancellations, further enhancing the practical applicability of our findings.
\label{section:conclusion}

\printbibliography[heading=subbibintoc]

\end{document}